\documentclass[10pt, a4paper]{article}

\usepackage[]{lrec-coling2024} 
 \usepackage{amsmath}

 \usepackage{times}
\usepackage{latexsym}
\usepackage{amsmath}
\usepackage{booktabs}
\usepackage{multirow}
\usepackage{siunitx}
\usepackage{graphicx}
\usepackage{float}
\usepackage{subcaption}
\usepackage{xcolor}

\newcommand{\myheart}{\textsuperscript{$\heartsuit$}}
\newcommand{\myspadesuit}{\textsuperscript{$\spadesuit$}}
\newcommand{\mydiamondsuit}{\textsuperscript{$\diamondsuit$}}

\title{Probing Large Language Models from A Human Behavioral Perspective}

\name{Xintong Wang\myspadesuit, Xiaoyu Li\myheart, Xingshan Li\mydiamondsuit, Chris Biemann\myspadesuit} 

\address{\myspadesuit Department of Informatics, Universität Hamburg \\ \myheart School of Computer Science and Technology, Beijing Institute of Technology \\ \mydiamondsuit Institute of Psychology, Chinese Academy of Sciences \\
         \{xintong.wang, chris.biemann\}@uni-hamburg.de, \\
         demo.xyli@gmail.com, lixs@psych.ac.cn\\
         }

\abstract{
Large Language Models (LLMs) have emerged as dominant foundational models in modern NLP. However, the understanding of their prediction processes and internal mechanisms, such as feed-forward networks (FFN) and multi-head self-attention (MHSA), remains largely unexplored. In this work, we probe LLMs from a human behavioral perspective, correlating values from LLMs with eye-tracking measures, which are widely recognized as meaningful indicators of human reading patterns. Our findings reveal that LLMs exhibit a similar prediction pattern with humans but distinct from that of Shallow Language Models (SLMs). Moreover, with the escalation of LLM layers from the middle layers, the correlation coefficients also increase in FFN and MHSA, indicating that the logits within FFN increasingly encapsulate word semantics suitable for predicting tokens from the vocabulary.
 \\ \newline  }

\begin{document}

\maketitleabstract

\section{Introduction}
Recent advancements in Large Language Models (LLMs) \cite{devlin2018bert, radford2019language, touvron2023llama, touvron2023llama2} have showcased their superior capabilities in language understanding, generation as well as zero-shot transferring. Despite their remarkable successes, issues such as the generation of hallucinated \cite{rawte2023survey} and toxic outputs \cite{leong2023self} have arisen, underscoring the importance of understanding the internal mechanisms and predictive behaviors of LLMs to develop models that are both powerful and reliable.

\begin{figure}[htbp]
  \centering
  \includegraphics[width=1.0\columnwidth]{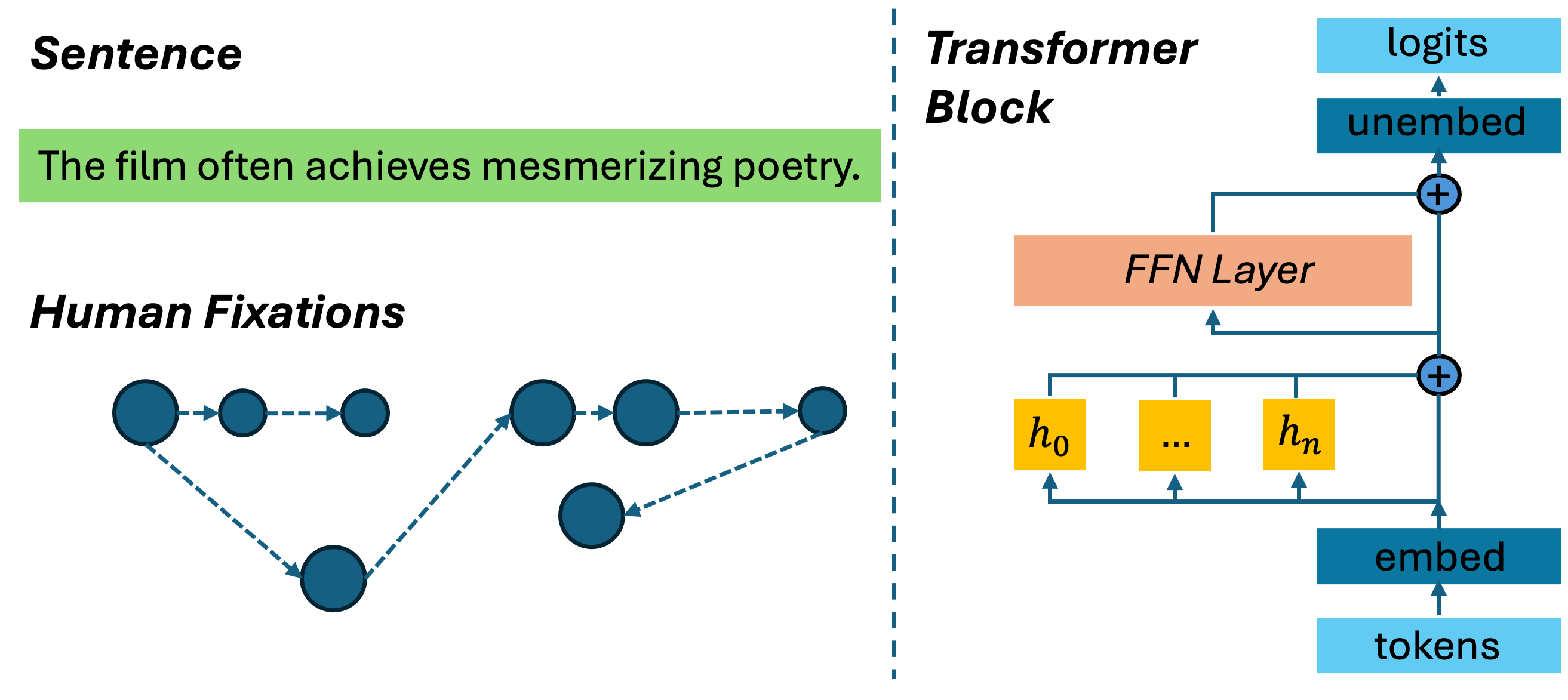}
  \caption{\textbf{Comparison of Human reading pattern and transformer block.} The left part shows the \textit{fixation patterns} of a human reader over a given sentence, while the right part demonstrates a transformer block including \textit{FFN layers and multi-head self-attention}. The \textcolor{blue}{blue dots} mark fixations on the corresponding words above; a wider diameter represents a longer fixation duration.}
  \label{fig:intro} 
\end{figure}

Research on LLM interpretation has emerged \cite{zhao2023explainability, wang2023knowledge}, focusing on dissecting the components of \textit{Feed-Forward Layers (FFN)} and \textit{Multi-Head Self-Attention (MHSA)}. \cite{geva-etal-2022-transformer} highlighted the role of FFN in LLMs, demonstrating how tokens are promoted by utilizing logits in the late layers for word prediction from a vocabulary. \cite{bills2023language} explored the activation of self-attention heads under varying prompts. Concurrently, cognition and psycholinguistic studies have documented various measures during human reading activities \cite{hollenstein2018zuco, hollenstein2019zuco, cop2017presenting, luke2018provo}, closely paralleling the processes observed in language models \cite{hofmann2022language}. As depicted in Figure~\ref{fig:intro}, the juxtaposition of \textit{human reading patterns} and a \textit{transformer block} illustrates the similarity in attention allocation—eye-tracking measurements for humans and FFN/MHSA values for LLMs—motivating our approach to probe LLMs from a human behavioral perspective.

Specifically, we investigate the \textbf{internal workings of FFN and MHSA} in LLMs, such as the GPT-2 model \cite{radford2019language}, by \textit{correlating eye-tracking fixations with LLM values}. Our findings reveal that LLMs, particularly in their \textbf{\textit{middle layers}}, increasingly mirror human attention patterns, focusing more on essential words. However, in contrast to humans who prioritize \textit{crucial content}, the \textbf{\textit{upper layers}} of LLMs refine context understanding, indicating a divergence in focus on \textit{less critical aspects}. This suggests that the outputs of FFN in the \textbf{\textit{upper layers}} can facilitate predictions beyond just the final layers, encouraging methods for efficient semantic editing \cite{wang2023knowledge}.

Furthermore, our comparison of \textbf{prediction behaviors} between LLMs and Shallow Language Models (SLMs) reveals that \textit{LLMs more closely resemble human predictive patterns}, where greater emphasis on significant words enhances the certainty of word predictions.

Our \textbf{contributions} are as follows:

\begin{itemize}
\item We conduct a detailed analysis of the internal mechanisms of FFN and MHSA in LLMs from a human behavioral perspective.
\item We juxtapose the word prediction processes of LLMs and SLMs, reinforcing the evidence that LLMs more closely align with human attention patterns, focusing on crucial words to enhance prediction certainty.
\end{itemize}
\section{Related Work}
\label{sec:append-how-prod}
\textbf{Human Behavior Measures}: Studies in cognition and psycholinguistics have deployed simultaneous eye-tracking and electroencephalography during natural and task-specific reading to comprehend human reading processes. Noteworthy datasets in this context include ZuCo 1.0 \cite{hollenstein2018zuco}, ZuCo 2.0 \cite{hollenstein2019zuco}, GECO \cite{cop2017presenting}, and Provo \cite{luke2018provo}. However, to the best of our knowledge, there is a paucity of work utilizing these datasets to probe LLMs and their internal mechanisms.

\textbf{Eye-movement Prediction}: A shared task at ACL 2021 \cite{hollenstein-etal-2021-cmcl} involved using language models for predicting eye-movement measures. In this shared task, models, including Boosting, MLP, and RoBERTa, displayed significant performance in this task. Besides, linguistic features proved crucial for achieving superior results \cite{bestgen2021last}. In this paper, we focus on employing eye-movement data for probing LLMs.

\section{Preliminary}
\textbf{Large language models (LLMs)} predominantly rely on the Transformer architecture \cite{vaswani2017attention}, composed of Transformer blocks acting as layers denoted by $l = 1, 2..., L$. As shown in Figure~\ref{fig:intro}, each Transformer block primarily consists of \textbf{multi-heads self-attention} and a \textbf{feed-forward network}. The motivation for the multi-head self-attention mechanism lies in its ability \emph{to extract various aspects of the sequence, with its capacity deepening with the increase of layers}. Concurrently, the FFN serves to \emph{output for the current layers and makes prediction over a vocabulary}.

More specifically, in layer $l$, the currently processed representation is denoted by $X_i^l$, and the output for FFN is computed as: 
\begin{equation}
    o_i^l=F F N^l\left(X_i^l\right),
\end{equation}
where $o_i^l$ denotes the output for the current FFN.

An updated representation $\tilde{x}_i^l$, is then achieved by adding $X_i^l$ and $o_i^l$. The updated representation, $\tilde{x}_i^l$, subsequently undergoes a self-attention process. Given the presence of multi-head self-attention in each layer, all the representations in each self-attention head are concatenated to serve as the input for the subsequent FFN layer, as illustrated below:
\begin{equation}
X_i^{l+1}=\operatorname{concatenate}\left(\text { Attention}^l\left(\tilde{x}_i^l\right)\right),
\end{equation}

In this work, we present empirical evidence understanding the function of multi-head self-attention and FFN layers by correlating their values with human behavioral data, eye-tracking measurements.

\section{Eye-tracking Measurements}

Human behavioral signals, such as \textbf{eye-tracking, fMRI, and EEG}, have been widely utilized in cognition and psycholinguistic studies. Among these signals, eye-tracking offers millisecond-precise recordings of gaze direction, illuminating the \textit{focus of attention during reading and comprehension}. This process bears resemblance to the operations within a \textbf{transformer block}, as depicted in Figure~\ref{fig:intro}. Thus, we employ \textbf{\textit{eye-tracking data}} to uncover the internal mechanics of the transformer architecture.

\begin{table}[htbp]
\centering
\tiny
\begin{tabular}{p{0.2\columnwidth}p{0.1\columnwidth}p{0.55\columnwidth}}
\toprule
\textbf{Eye-movement Measures} & \textbf{Abbrev.} & \textbf{Definition} \\
\midrule
Gaze duration & GD & The sum of all fixations on the current word in the first-pass reading before the eye moves out of the word \\
Total reading time & TRT & The sum of all fixation durations on the current word, including regressions \\
First fixation duration & FFD & The duration of the first fixation on the prevailing word \\
Single fixation duration & SFD & The duration of the first and only fixation on the current word \\
Go-past time & GPT & The sum of all fixations prior to progressing to the right of the current word, including regressions to previous words that originated from the current word \\
\bottomrule
\end{tabular}
\caption{\textbf{Definition of Five Eye-tracking Measures}: Gaze Duration (GD), Total Reading Time (TRT), First Fixation Duration (FFD), Single Fixation Duration (SFD), and Go-Past Time (GPT).}
\label{tab:Table eye} 
\end{table}

In our study, we establish correlations between metrics derived from \textit{multi-head self-attention (MHSA), feed-forward neural (FFN) layers}, and \textbf{five specific eye-tracking measurements}: \textit{Gaze Duration (GD), Total Reading Time (TRT), First Fixation Duration (FFD), Single Fixation Duration (SFD), and Go-Past Time (GPT).} Each of these metrics offers unique insights into the human reading process. For instance, Gaze Duration (GD) refers to the cumulative duration of all fixations on a given word during initial reading before moving to the next word, with \textit{longer durations indicating the word's significance}. Similarly, Total Reading Time (TRT) encompasses all fixation durations on a word, including regressions, indicating that \textit{readers may revisit a word multiple times to refine their understanding}. The detailed meanings of these eye-tracking measures can be found in Table~\ref{tab:Table eye}.

By leveraging these interpretable eye-tracking metrics, we aim to probe LLMs by correlating their values with those observed in multi-head attention and FFN layers.

\section{Experiments} 


\subsection{Experimental Settings}
\textbf{Language Models:} For our investigation, we utilized a pre-trained GPT-2 model (\textit{base}) from HuggingFace, focusing on analyzing the internal mechanisms of FFN and multi-head self-attention mechanisms due to its \textbf{simplicity and general applicability}. We posit that our probing method is adaptable and can be extended to other, more advanced open-source LLMs such as LLaMA \cite{touvron2023llama} and Qwen \cite{bai2023qwen}, among others. Additionally, we broaden our analysis to include \textbf{Shallow Language Models (SLMs)} like N-Gram language models \cite{pauls2011faster}, Recurrent Neural Networks (RNNs), Gated Recurrent Units (GRUs), Long Short-Term Memory (LSTM) networks \cite{sherstinsky2020fundamentals}, and a recently enhanced RNN variant, the RWKV-V4 model \cite{peng2023rwkv}, to conduct a comprehensive comparison of prediction probabilities. For the training of SLMs, we employ the WikiText-103 dataset.

\textbf{Eye-tracking Data:} For human behavioral data, we utilize the ZuCo 2.0 dataset \cite{hollenstein2019zuco}, which contains concurrent eye-tracking records captured during two types of reading activities: \textit{natural reading (NR) and task-specific reading (TSR)}. This dataset is notably comprehensive, comprising 730 English sentences, split into 349 sentences read under normal conditions and 390 sentences read under a task-specific paradigm. Eye-tracking data from 18 participants were recorded during both NR and TSR activities. We conducted word prediction experiments using various language models on sentences from the ZuCo 2.0 dataset to then analyze the correlation patterns between human reading behaviors and language model predictions.

\textbf{Correlation Metrics and Evaluation:} Following previous studies \cite{eberle2022transformer} on analyzing the prediction behavior of LLMs, we also employ three prevalent correlation metrics: Pearson \cite{freedman2007statistics}, Spearman \cite{caruso1997empirical}, and Kendall \cite{abdi2007kendall}, to investigate the relationship between values derived from LLMs and human behavioral measures. Despite minor differences, we find these correlation metrics yield similar results. Among them, Spearman exhibits superior robustness when compared to Pearson and Kendall. Unless stated otherwise, experimental results are reported using Spearman analysis. \textit{Given that larger fixations, as indicated by various eye-tracking measures, signify the importance of the current word, \textbf{a stronger correlation implies that LLMs also allocate more attention to the processed word.}}

\begin{figure}[htbp]
  \centering
  \includegraphics[width=1.0\columnwidth]{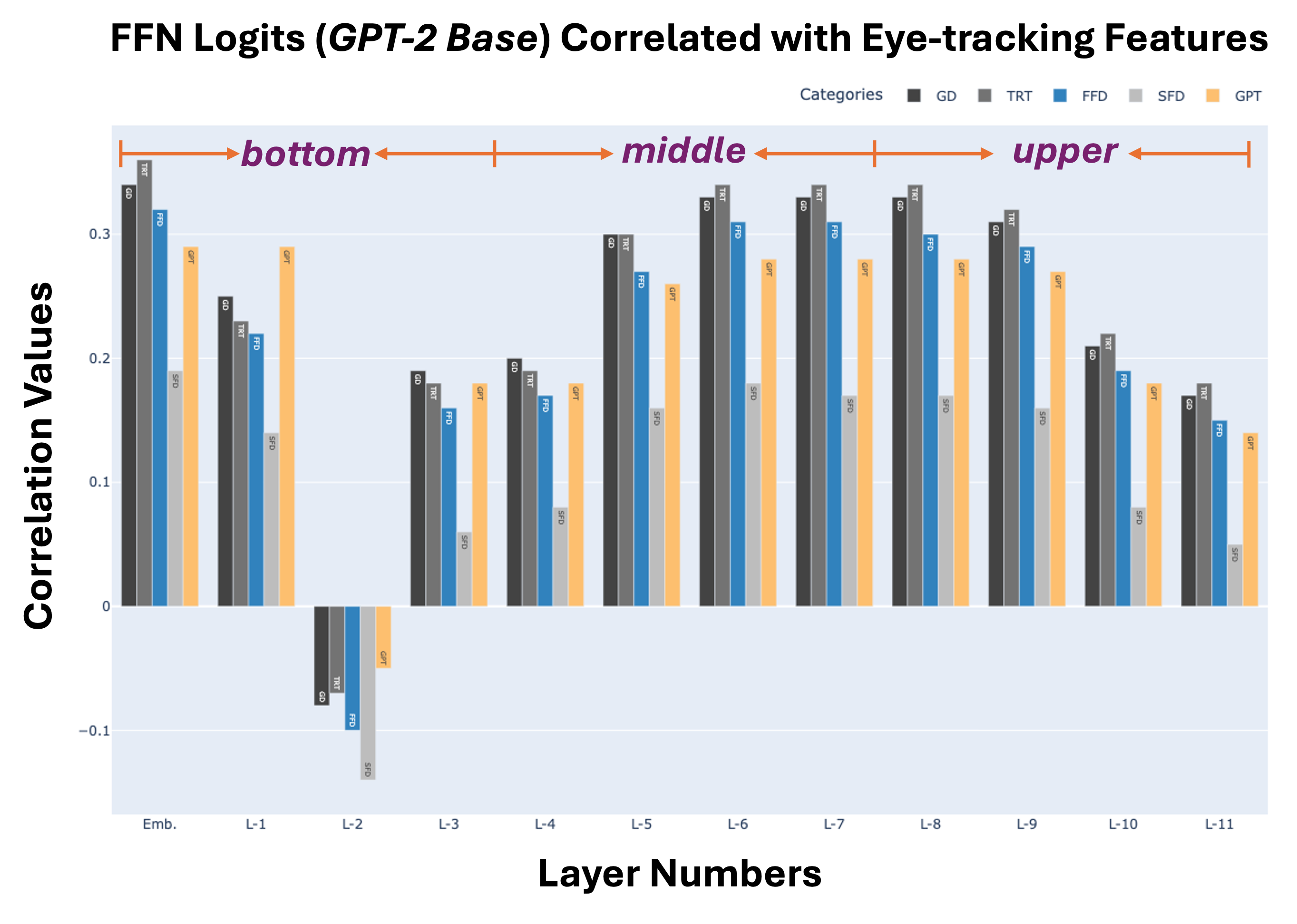}
  \caption{\textbf{FFN Correlation Values.} FFN values through layers in GPT-2 \textit{base} Correlated with five different eye-tracking features in three groups: bottom, middle, and upper. (Significant at $p \textless 0.05$)}
  \label{fig:ffn} 
\end{figure}

\begin{figure*}[htbp] 
  \centering
  \includegraphics[width=1\textwidth]{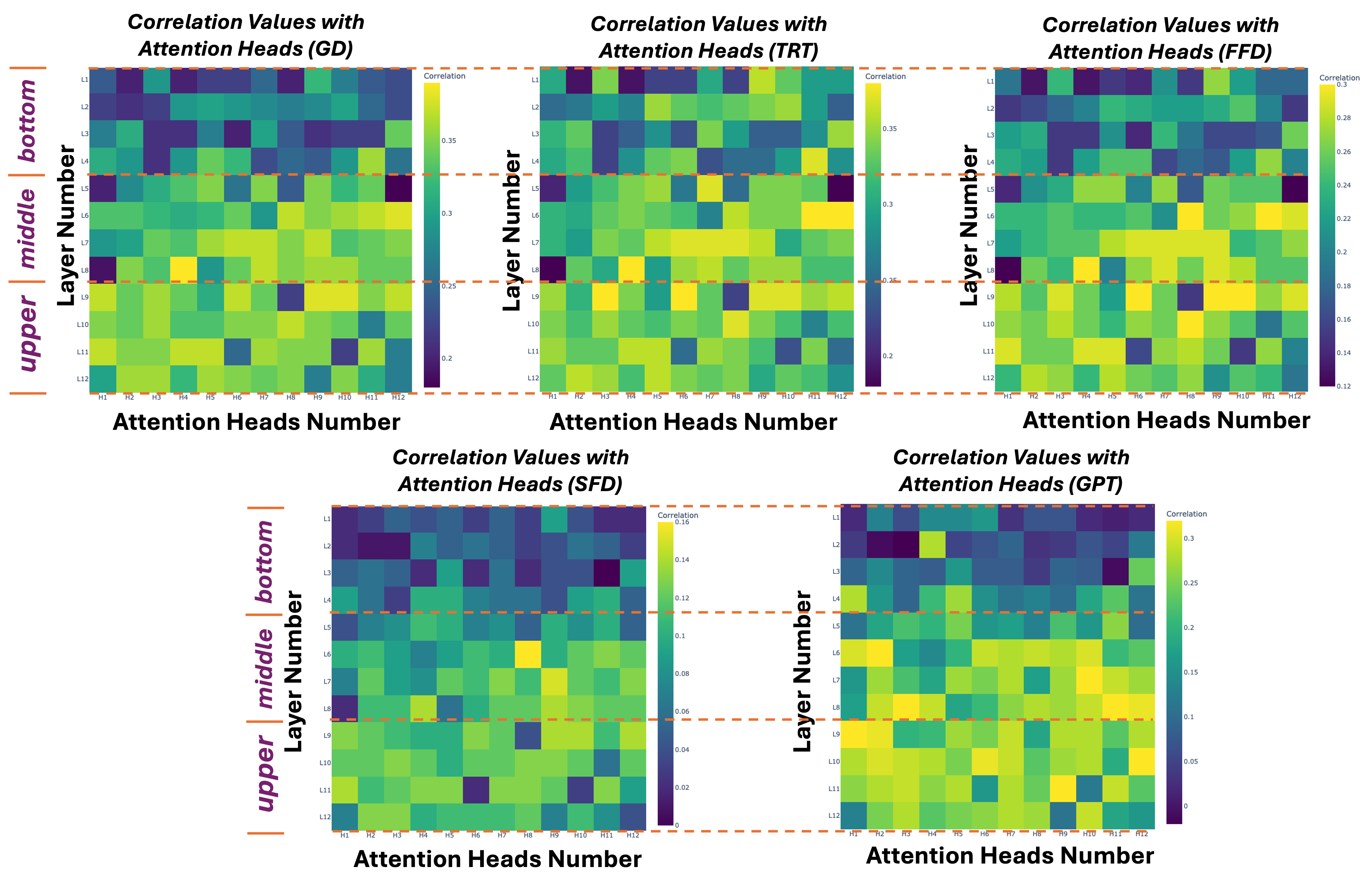} 
  \caption{\textbf{Attention Heads Correlated Values with Eye-tracking Measurements through Layers Results.} \textcolor{lime}{\textit{\textbf{Lighter and larger values}}} signify stronger correlations.}
  \label{fig:mha} 
\end{figure*}

\subsection{FFN Correlation Analysis}
We examine the functions of the FFN within GPT-2. To elucidate our findings, we categorize the 12 layers of GPT-2 (\textit{base}) into three groups: \textbf{bottom ($l_{1} \rightarrow l_{4}$), middle  ($l_{5} \rightarrow l_{8}$), and upper  ($l_{9} \rightarrow l_{12}$)}. As illustrated in Figure~\ref{fig:ffn}, the bottom most layers show a direct correlation between the embedding of input tokens and human reading fixations. This suggests that humans require more time to comprehend critical tokens that are also reflected in the embeddings of LLMs. This correlation diminishes as we ascend through the layers, with the topmost layer of the bottom group (Layer 3) indicating a divergence in processing tokens from human behavior; the FFN at this level begins to process tokens yet in a manner distinct from human reading patterns.


Progressing to the middle layers, the correlation coefficients initially increase and then stabilize, peaking at Layer 6. This pattern suggests that the FFN in these middle layers starts to show similar human fixation behaviors, indicating that the logits within FFN increasingly encapsulate word semantics suitable for predicting tokens from the vocabulary.

Intriguingly, in the upper layers, we observe a decline in correlation values. We hypothesize that at this stage, the LLM begins to incorporate less critical words within sentences into its consideration, diverging from human intuition, which tends to focus on the most crucial aspects of the context and disregard less important information.


\subsection{Multi-head Self-attention Correlation}

Figure~\ref{fig:mha} presents heatmaps that illustrate the correlation between the values of 12 self-attention heads across 12 distinct layers and human behavioral data; where lighter and larger values signify stronger correlations. Similar to our FFN analysis, we categorized the 12 layers into three groups: \textbf{bottom, middle, and upper}. The bottom group exhibits a weaker correlation with human fixations, indicating that while self-attention mechanisms begin to process tokens at this stage, they do so differently from human behavior.

As we ascend through the middle and upper groups, we observe an increase in correlation across different layers and attention heads with human fixations. This pattern suggests that, in these layers, LLMs begin to align more closely with human patterns, especially in focusing on important contextual tokens. Notably, unlike in the FFN analysis, we did not observe a decrease in multi-head attention correlation values in the upper layers. This difference implies that the comprehension capabilities of LLMs are progressively refined up to the final layer, enabling more diverse and accurate word predictions compared to human reading patterns.

Furthermore, among the five eye-tracking measures analyzed, Gaze Duration (GD), Total Reading Time (TRT), First Fixation Duration (FFD), and Go-Past Time (GPT) demonstrate stronger correlations, whereas Single Fixation Duration (SFD) shows a weaker correlation. Given that SFD represents the first and only fixation on a current word—suggesting lesser importance—while GD, TRT, FFD, and GPT include regressions on significant words, this discrepancy explains why LLMs also prioritize these important words.


\subsection{Prediction Probability Correlation}
We further analyze word prediction probability behaviorals in LLMs and our investigation into the correlation of word prediction probabilities reveals distinct behaviors between Large Language Models (LLMs) and Shallow Language Models (SLMs). For this analysis, we employed two reading tasks: task-specific reading (TSR) and natural reading (NR). The TSR task encompassed 5335 words for prediction analysis, while the NR task included 5329 words. Our findings, detailed in Table~\ref{tab: prob}, are divided into two parts: the upper section presents the correlation outcomes for the TSR task, and the lower section for the NR task.

Overall, SLMs exhibit a notable and consistent \textbf{negative correlation} in both the TSR and NR tasks. This trend suggests that SLMs tend to assign higher prediction probabilities with fewer fixations on critical words, thereby increasing the uncertainty of word predictions. In contrast, LLMs, exemplified by GPT-2, demonstrate a significant and \textbf{positive correlation} in both tasks. This positive correlation indicates that LLMs exhibit a prediction pattern akin to human behavior, where increased attention to crucial words leads to more confident predictions.

Though the aforementioned conclusions are consistent for both the TSR and NR tasks, it is noteworthy that the correlation values for the NR task are consistently higher than those for the TSR task. We hypothesize that during task-specific readings, humans are guided by specific clues to identify and concentrate on words that are pertinent to the task at hand. Consequently, our word prediction analysis across different LMs aligns more closely with the process in NR.

\begin{table}[H]  
\centering
\resizebox{\columnwidth}{!}{%
\begin{tabular}{l*{5}{cS[table-format=-1.2]c}}
\toprule
\multirow{2}{*}{\textbf{Model}} & \multicolumn{15}{c}{\textbf{Eye-tracking Measures}} \\
\cmidrule(lr){2-16}
& \multicolumn{3}{c}{GD} & \multicolumn{3}{c}{TRT} & \multicolumn{3}{c}{FFD} & \multicolumn{3}{c}{SFD} & \multicolumn{3}{c}{GPT} \\
\midrule
\multicolumn{16}{c}{Task-specific Reading} \\
\midrule
N-Gram & & -0.26 & & & -0.25 & & & -0.23 & & & -0.15 & & & -0.23 & \\
RNN & & -0.44 & & & -0.43 & & & -0.41 & & & -0.28 & & & -0.40 & \\
GRU & & \textcolor{blue}{-0.46} & & & \textcolor{blue}{-0.45} & & & \textcolor{blue}{-0.43} & & & \textcolor{blue}{-0.30} & & & \textcolor{blue}{-0.43} & \\
LSTM & & -0.42 & & & -0.41 & & & -0.39 & & & -0.26 & & & -0.39 & \\
RWKV & & -0.39 & & & -0.40 & & & -0.40 & & & -0.27 & & & -0.33 & \\
GPT-2 & & \textcolor{red}{0.23} & & & \textcolor{red}{0.21} & & & \textcolor{red}{0.20} & & & \textcolor{red}{0.12} & & & \textcolor{red}{0.28} & \\
\midrule
\multicolumn{16}{c}{Natural Reading} \\
\midrule
N-Gram & & -0.33 & & & -0.33 & & & -0.31 & & & -0.15 & & & -0.29 & \\
RNN & & -0.52 & & & -0.51 & & & -0.50 & & & -0.26 & & & -0.46 & \\
GRU & &  \textcolor{blue}{-0.54} & & & \textcolor{blue}{-0.53} & & & \textcolor{blue}{-0.52} & & & \textcolor{blue}{-0.29} & & & \textcolor{blue}{-0.48} & \\
LSTM & & -0.52 & & & -0.50 & & & -0.49 & & & -0.26 & & & -0.46 & \\
RWKV & & -0.39 & & & -0.39 & & & -0.38 & & & -0.19 & & & -0.28 & \\
GPT-2 & & \textcolor{red}{0.33} & & & \textcolor{red}{0.30} & & & \textcolor{red}{0.30} & & & \textcolor{red}{0.14} & & & \textcolor{red}{0.37} & \\
\bottomrule
\end{tabular}
}%
\caption{\textbf{Prediction Probability Correlation Results} using Spearman correlation metric. The numbers in \textcolor{blue}{blue} mean the significant negative correlation, while the \textcolor{red}{red} represent the positive correlation. (Significant at $p\textless0.05$)}
\label{tab: prob}
\end{table}

\section{Conclusion}
In this work, we probe LLMs through human behavior, specifically employing eye-tracking measurements to dissect the internal workings of LLMs, including the feed-forward layers and multi-head attention. Our findings reveal a similarity between LLMs and humans on word prediction: both exhibit a tendency where heightened attention to pivotal words results in more confident predictions. Our analysis further delineates that feed-forward networks begin to align with human fixation patterns starting from the middle layers, leveraging upper layers to broaden the contextual understanding. Our probing approach stands out for its interpretability from human reading indicators and paves the way for the development of LLMs that are not only reliable but also imbued with a greater degree of trustworthiness.

\section{Acknowledgements}
This research was funded by the German Research Foundation DFG Transregio SFB 169: Crossmodal Learning: Adaptivity, Prediction, and Interaction.


\nocite{*}
\section{Bibliographical References}\label{sec:reference}

\bibliographystyle{lrec-coling2024-natbib}
\bibliography{lrec-coling2024-example}


\section{Ethical Considerations}
The eye-tracking data employed in this study, derived from the ZuCo 2.0 dataset, are publicly accessible and adhere to established ethical protocols.

\end{document}